\documentclass{article}

\usepackage[preprint]{neurips_2022}

\usepackage[utf8]{inputenc} 
\usepackage[T1]{fontenc}    
\usepackage{hyperref}       
\usepackage{url}            
\usepackage{booktabs}       
\usepackage{amsfonts}       
\usepackage{nicefrac}       
\usepackage{microtype}      
\usepackage{xcolor}         

\usepackage{graphicx}
\usepackage{natbib}
\usepackage{doi}
\usepackage{latexsym}
\usepackage{amsmath}
\usepackage{amssymb}
\usepackage{multirow}
\usepackage{pgfplots}
\pgfplotsset{compat=1.17}

\usepackage[ruled]{algorithm2e}
\SetKwComment{Comment}{/* }{*/ }
\SetKwInOut{Parameter}{parameter}
\usepackage{fourier} 
\usepackage{array}
\usepackage{makecell}

\usepackage{newfloat}
\usepackage{listings}
\usepackage{xcolor}
\usepackage{amsthm}
\usepackage[english]{babel}
\theoremstyle{definition}
\newtheorem{definition}{Definition}[section]
\newtheorem{proposition}{Proposition}[section]
\newtheorem{theorem}{Theorem}[section]

\newtheorem{remark}{Remark}[section]

\title{Additive Logistic Mechanism for Privacy-Preserving Self-Supervised Learning}

\author{%
  Yunhao Yang \\
  Department of Computer Science\\
  University of Texas at Austin\\
  Austin, TX 78705 \\
  \texttt{yunhaoyang234@utexas.edu} \\
   \And
   Parham Gohari \\
   Department of Electrical and Computer Engineering\\
   University of Texas at Austin\\
   Austin, TX 78705 \\
   \texttt{pgohari@utexas.edu} \\
   \AND
   Ufuk Topcu\\
   Oden Institute for Computational Engineering and Sciences\\
   University of Texas at Austin\\
   Austin, TX 78705 \\
   \texttt{utopcu@utexas.edu} \\
}

\begin{document}

\maketitle

\begin{abstract}
    We study the privacy risks that are associated with training a neural network's weights with self-supervised learning algorithms.
    Through empirical evidence, we show that the fine-tuning stage, in which the network weights are updated with an informative and often private dataset, is vulnerable to privacy attacks.
    To address the vulnerabilities, we design a post-training privacy-protection algorithm that adds noise to the fine-tuned weights and propose a novel differential privacy mechanism that samples noise from the logistic distribution.
    Compared to the two conventional additive noise mechanisms, namely the Laplace and the Gaussian mechanisms, the proposed mechanism uses a bell-shaped distribution that resembles the distribution of the Gaussian mechanism, and it satisfies pure $\epsilon$-differential privacy similar to the Laplace mechanism.
    We apply membership inference attacks on both unprotected and protected models to quantify the trade-off between the models' privacy and performance.
    We show that the proposed protection algorithm can effectively reduce the attack accuracy to roughly 50\%—equivalent to random guessing—while maintaining a performance loss below 5\%.
\end{abstract}
\section{Introduction}

Machine learning algorithms, and their outcomes, are known to be vulnerable to privacy attacks. Attackers may, for example, infer the size of the training data \citep{dalenius1977towards, Dwork2010OnTD}, individuals’ information in the training dataset \citep{Shokri_2017}, and parameters of machine learning models \citep{Ateniese2015HackingSM, Ohrimenko2016ObliviousMM} by only observing the inputs and outputs of these models.

We focus on the vulnerability due to privacy attacks in the \textit{fine-tuning stage of self-supervised learning algorithms}. Many self-supervised models are first pre-trained using public databases and then fine-tuned using private datasets for specific purposes \citep{chen2020simple, Devlin2019BERTPO}. Our first contribution is an empirical demonstration of this vulnerability. To this end, we devise a membership inference attack that obtains an accuracy of 71\% on the Cifar 100 dataset \citep{cifar} and 61\% on the STL 10 dataset \citep{deng2009imagenet}. Note that such attack accuracy is significantly higher than the accuracy of random guessing, hence protections against privacy attacks are necessary.

Federated learning \citep{konen2015federated} and data encryption \citep{encryption} are among the commonly used protection mechanisms against privacy attacks to machine learning algorithms. On the other hand, they have limited applicability for self-supervised learning. Fine-tuning datasets are often small; therefore, partitioning of the dataset as in federated learning is not practical or may result in a loss of convergence. Data encryption, when applied only on fine-tuning datasets, may introduce inconsistency between pre-training and fine-tuning datasets, again resulting in a loss of convergence. 

As for the second contribution, we develop a new differential privacy mechanism for protecting self-supervised learning algorithms against privacy attacks.
This new mechanism, which we call \textit{the additive logistic mechanism}, adds small amounts of statistical noise sampled from a logistic distribution to conceal the contributions of individual parties in the fine-tuning dataset. Besides the additive logistic mechanism, several other additive noise mechanisms have been developed to protect the privacy of machine learning models \citep{Abadi2016DeepLW, ji2014differential}. Two of the most commonly used additive noise mechanisms are the Gaussian mechanism \citep{gaussian-mechanism} and the Laplace mechanism \citep{laplace-mechanism}, drawing noise from a Gaussian or Laplace distribution.
The proposed additive logistic mechanism uses a bell-shaped distribution that resembles the distribution of the Gaussian mechanism, and it satisfies pure $\epsilon$-differential privacy similar to the Laplace mechanism.
We later empirically show how the proposed additive logistic mechanism outperforms these two in protecting self-supervised learning algorithms.

While the new mechanism is broadly applicable, as the third contribution, we show that it can be adopted for the protection of self-supervised neural networks. Specifically, we develop a post-training protection algorithm that deploys the additive logistic mechanism to the self-supervised neural network. This protection algorithm provides privacy to the network by adding noise to its fine-tuning layer-weights. We theoretically prove that the protection algorithm satisfies differential privacy and compute the privacy budget in terms of the parameters of the logistic distribution. This post-training protection applies to already fine-tuned networks, hence, it avoids the aforementioned loss of convergence. 

As for the final contribution, we illustrate the privacy-performance trade-off in the protected self-supervised network. Adding noise to the layer-weights degrades the performance of the self-supervised neural network. We show the negative correlation between the privacy and performance of the self-supervised network and indicate the additive logistic mechanism can achieve a high privacy level- $\epsilon < 1$- with a performance loss below 10\%.

In the empirical analysis, we perform membership inference attacks to show that the protection algorithm can effectively reduce the attack accuracy to approximately random guessing. Then, we evaluate and compare the privacy-performance trade-offs of the three mechanisms- the Gaussian mechanism, the Laplace mechanism, and the additive logistic mechanism- to show the performance loss of the additive logistic mechanism is lower than the other two under the same privacy budget $\epsilon$. Additionally, we compare the protection strength against membership inference attacks of these three mechanisms. We show that the proposed additive logistic mechanism achieves the strongest protection against membership inference attacks while having the lowest loss in performance among all three mechanisms.

The privacy attacks and protections on self-supervised learning we have addressed in this paper are notable because self-supervised learning has contributed to the state of the arts in multiple fields and many of them are irreplaceable. Self-supervised learning algorithms have been used in language generation \citep{Devlin2019BERTPO, Radford2018ImprovingLU, mikolov2013efficient}, image classification \citep{chen2020simple, contrastive}, image and video enhancing or super-resolution \citep{zhang2021idr, laine2019highquality, Huang2021Neighbor2NeighborSD}, robotics \citep{Endrawis2021EfficientSD}, etc. These algorithms are the state of the arts in their own fields and there is no substitution that can achieve the same performances yet. Therefore, it is important to notice the privacy vulnerabilities of self-supervised learning and address these vulnerabilities.

\section{Related Works} 
K-anonymity \citep{k-anonymity}, plausible deniability \citep{plausible-deniability}, and data obfuscation \citep{obfuscate} are pre-training protections that add noise sampled from a Gaussian distribution to the raw training data before feeding them into the networks. Adding noise to the raw data is inefficient when dealing with large-scale data records, such as images and videos. The protection algorithm that deploys the additive logistic mechanism is operating on the weights of the networks, hence not concerning the scale of data records.

DP-SGD \citep{Abadi2016DeepLW} clips the gradient and adds noise to the gradient during training, which significantly increases computation complexity. Deploying DP-SGD with a defined noise level required retraining the network. By contrast, we can adjust the noise level of the post-training protection without retraining the network.
\cite{he2021quantifying} develops a privacy-preserving contrastive learning mechanism, Talos, relying on adversarial training. Talos is in-training protection operated in the pretraining stage, which slows down the pretraining stage. The post-training protection algorithm does not increase the training time in any stage.

Existing post-training protections \citep{yang2021privacy, lu2022differentially} add noises to the neurons of the network. The privacy budget of such protections increases when the size of queries grows. The proposed protection algorithm that deploys the additive logistic mechanism adds noises to the trainable weights, hence will not be affected by the data scale.

\section{Preliminary}
\label{sec:prelim}

\paragraph{Self-Supervised Learning}
is a machine learning technique that can learn features from unlabeled training data. We focus on self-supervised learning algorithms applied to neural networks. These algorithms typically train neural networks in two stages \citep{ssl1, ssl2}: the \textit{pre-training stage} and \textit{the fine-tuning stage}.
In the pre-training stage, we initialize the network weights by solving a separate machine learning task using ``pseudo-labels". Pseudo-labels are automatically generated labels from a given \textit{pre-training dataset}. For example, we can rotate a training image and consider the rotation angle as a pseudo-label. We refer to the resulting machine learning model in this stage as the \textit{base encoder}.
In the fine-tuning stage, we fine-tune the base encoder for the main machine learning task that could be supervised, unsupervised, or reinforcement learning.
In particular, attach a \textit{projection head}, which is basically a collection of additional layers of nodes, to the base encoder's output layer. In the fine-tuning stage, it is the projection weights that are updated. We refer to the training dataset that we use in this stage as the \textit{fine-tuning dataset}. The base encoder preserves the features of the input data while mapping the data to the output space. The projection head uses these features to solve the actual task.

\paragraph{Membership Inference Attack (MIA)}
is one of the well-known privacy attacks that can be applied to neural networks to infer whether a selected data record belongs to the training dataset of the networks \citep{jia202110, yeom2017privacy, Salem2019}. Researchers have developed several effective MIA techniques and have evaluated the vulnerability of various neural networks to MIAs \citep{Shokri_2017, li2020membership, Liu2021MLDoctorHR}. We follow the framework of \textsc{shadow models} \citep{Shokri_2017} to perform black-box MIAs, in which the attacker can only access the input and output layer of the networks.

A shadow model must mimic the behavior of the victim model using a known training dataset, where each record in the dataset consists of a sample and a ground-truth label. Then, we feed a record into the shadow model to obtain an output and concatenate the output with its label to form an output-label pair. We train a binary classifier that takes the output-label pair as input to determine if the record has been used in training of the shadow model. We describe the detailed implementations in the Appendix.

\paragraph{Differential Privacy}
is a concept for publicly communicating messages about a dataset by describing the patterns of groups within the dataset while concealing the information of individual records in the dataset. The work \citep{laplace-mechanism} has introduced the concept of $\epsilon$-differential privacy. The value of $\epsilon$ evaluates the privacy budget of an algorithm by measuring the contribution of each individual records to the algorithm.
\begin{definition}[\textbf{\textsc{$\epsilon$-differential privacy}}]
    \label{def:DP}
    \citep{gaussian-mechanism}.
    Let $\epsilon$ be a positive real number and $M$ be a randomized algorithm that takes a dataset as input. Let $\mathcal{Y}$ be the image of $M$. The algorithm $M$ is $\epsilon$-differentially private if, for all datasets $D_1$ and $D_2$ that differ in a single element, and all $R \subseteq \mathcal{Y}$:
    \begin{equation}
        \mathbb{P}[M(D_1) \in R] \le \exp (\epsilon) \mathbb{P}[M(D_2) \in R].
    \end{equation}
    
    Derived from the definition of $\epsilon$-differential privacy, an \textbf{\textsc{$(\epsilon, \delta)$-differential privacy}} must satisfy the following:
    \begin{equation}
        \mathbb{P}[M(D_1) \in R] \le \exp (\epsilon) \mathbb{P}[M(D_2) \in R] + \delta,
    \end{equation}
    where $\delta$ captures the probability that $\epsilon$-differential privacy fails.
\end{definition}

We use $\epsilon$ to quantify the privacy budget of the randomized algorithm. A lower $\epsilon$ indicates the algorithm maintains better privacy. If $\epsilon < 1$, then we can say that the algorithm maintains high privacy. An $\epsilon$ between 1 and 10 is a reasonable privacy level; and if $\epsilon > 10$ then we would consider this algorithm privacy risky \citep{gaussian-mechanism}.

\section{Additive Logistic Mechanism}

In this section, we introduce a new differentially private mechanism called additive logistic mechanism. We introduce the mechanism in Definition \ref{def:alm}.

\begin{definition}
    \label{def:alm}
    Let $f: \mathcal{X} \mapsto \mathcal{Y}$ be a fixed function and $\mathcal{Y}$ be a space in which addition is properly defined.
    The \textbf{\textsc{additive logistic mechanism (ALM)}} $M_{AL}$ takes the function $f$, a query point $x \in \mathcal{X}$, and a distribution parameter $s \in \mathbb{R}$ as inputs and returns a noisy output:
    \begin{center}
        $M_{AL}(x, f, s) = f(x) + Log(s)$,
    \end{center}
    where $Log(s)$ is an $n$-dimensional vector $\Vec{v}=\{v_1,...,v_n\} \in \mathbb{R}^n$ and each $v_i \in \Vec{v}$ is sampled from a logistic distribution with the location parameter $\mu = 0$ and the scale parameter $s$. The probability density function (PDF) of the logistic distribution is as follows:
    \begin{equation}
    \label{eq:pdf}
        p(x; s, \mu) = \frac{\exp(-(x - \mu)/s)}{s(1 + \exp(-(x - \mu)/s))^2}.
    \end{equation}
\end{definition}

\subsection{Differentially Private Additive Logistic Mechanism}
\label{sec:alm}
We claim that the ALM is $\epsilon$-differentially private and express the privacy budget $\epsilon$ in terms of the parameter $s$ (in Equation \ref{eq:pdf}). We begin the claim by introducing the concept of \textit{sensitivity}. When we take a function $f$ and feed a set of inputs to $f$, individual inputs may impact the output of $f$ differently. Sensitivity $\Delta_f$ is a measurement of the extend to which an individual component impacts the overall outcome of $f$ \citep{laplace-mechanism}. We formally define sensitivity as follows:
\begin{definition}
    \label{def: sensitivity}
    Let a function $f:\mathcal{X}\mapsto \mathcal{Y}$ be a mapping from a dataset domain $\mathcal{X}$ to a normed space $\mathcal{Y}$. Then, the sensitivity of the function $f$, denoted $\Delta_f$, is
    \begin{equation}
        \Delta_f := \sup_{D,D'} \|f(D) - f(D')\|,
    \end{equation}
    where $\|\cdot\|$ is the norm operator and $D$ and $D'$ are any two \textit{adjacent} datasets and the notion of adjacency is formally defined as follows:
\end{definition}

\begin{definition}
    \label{def:adjacent}
    Let $f$ be a given function. Define two datasets $D$ and $D'$, both in the input domain of $f$, \textit{adjacent} if the number of entries in which the two datasets hold different values is at most one.
\end{definition}

Once we compute the sensitivity $\Delta_f$, we can set the parameter $s$ of the ALM for a given privacy budget $\epsilon$.

\begin{theorem}
\label{thm:epsilon}
    Let $f$ be a given function and $\Delta_f$ be its sensitivity. For a given $\epsilon>0$, the additive logistic mechanism $M_{AL} (x, f, s)$ is $\epsilon$-differentially private for $\epsilon = \frac{\Delta_f}{s}$.
\end{theorem}

\begin{remark}
\label{remark:1-norm}
    Let $f: \mathcal{X} \mapsto \mathbb{R}^n$ be a function whose outputs are vector-valued; Theorem \ref{thm:epsilon} still holds. In this case, the sensitivity $\Delta_f$ is computed with respect to the 1-norm distance, denoted \textit{1-norm sensitivity}.
\end{remark}

\subsection{Existing Additive Noise Mechanisms for Differential Privacy}
We now review two of the most well-known additive noise mechanisms, the Laplace mechanism $M_{Lap}$ \citep{laplace-mechanism} and the Gaussian mechanism $M_{Gauss}$ \citep{gaussian-mechanism}. We use these two mechanisms as benchmarks and compare them with the additive logistic mechanism that we proposed.
The Gaussian mechanism and the Laplace mechanism have been proved to be differentially private and have been widely utilized in privacy protection algorithms \citep{Abadi2016DeepLW, yang2021privacy, lu2022differentially}. We now state the definitions of the two mechanisms.
\begin{remark}
    \label{remark:anm}
    The Laplace mechanism adds noise drawn from a Laplace distribution:
    \begin{center}
        $M_{Lap}(x, f, \epsilon) = f(x) + Lap(\mu=0, b=\frac{\Delta_f}{\epsilon})$
    \end{center}
    and is $\epsilon$-differentially private. $\Delta_f$ is the 1-norm sensitivity of $f$.
    
    The Gaussian mechanism adds noise drawn from a Gaussian distribution:
    \begin{center}
        $M_{Gauss}(x, f, \epsilon, \delta) = f(x) + \mathcal{N} \left( \mu=0, \sigma^2 = \frac{2 \ln(1.25/\delta) \cdot (\Delta_f)^2}{\epsilon^2} \right)$
    \end{center}
    and is $(\epsilon, \delta)$-differentially private \citep{10.1007/11761679_29}, where $\delta$ is typically less than $\frac{1}{N}$ and $N$ is the dataset size. $\Delta_f$ is the sensitivity of $f$ computed in the 2-norm distance, denoted \textit{2-norm sensitivity}.
\end{remark}

\subsection{Differential Privacy for Machine Learning Algorithms}
\label{sec:app}
We refer to a machine learning algorithm as an algorithm that is used to optimize the weights of a neural network using a given training dataset. A machine learning algorithm $\mathcal{A_H}$ takes in as input a set of hyper parameters $\mathcal{H}$ and a training dataset and outputs a set of weights in the real coordinate space.
Therefore, we claim that Theorem \ref{thm:epsilon} can be used in the context of machine learning algorithms that are used to train neural networks.


We deploy a randomized mechanism in the machine learning algorithm to add noise to the outputs of the algorithm. The randomized mechanism helps the algorithm satisfy differential privacy. Hence we can construct an ad-hoc \textit{differentially private machine learning algorithm}.

\begin{definition}
    \label{prop:DP}
    [\textbf{\textsc{Differentially Private Machine Learning Algorithm}}].
    
    Let $\mathcal{H}$ be a set of hyper-parameters and $\mathcal{A_H}: \mathcal{X} \mapsto \mathbb{R}^n$ be a machine learning algorithm that maps any training dataset $D \in \mathcal{X}$ to a vector of network weights.
    Let $M:\mathbb{R}^n \mapsto \mathbb{R}^n$ be a randomized mechanism, $D$ and $D'$ be two adjacent training datasets.
    We say $M(\mathcal{A_H})$ is $(\epsilon, \delta)$-differnetially private if, for all $R\subseteq \mathbb{R}^n$, all adjacent $D$ and $D'$, and all $x\in \mathcal X$,
    \begin{equation}
    \label{eq:diff privacy}
        \mathrm{P}\left[M(\mathcal{A_H}(D)) \in R \right] \le
        \exp(\epsilon) \cdot \mathrm{P}\left[M(\mathcal{A_H}(D')) \in R \right] + \delta.
    \end{equation}
    If $\delta = 0$, we call the algorithm $M(\mathcal{A_H}): \mathcal{X} \mapsto \mathbb{R}^n$ an \textit{$\epsilon$-differentially private machine learning algorithm}.
\end{definition}

Feeding two adjacent training datasets $D$ and $D'$ results in two different sets of network weights being trained. The sensitivity establishes an upper bound on the difference between these weights for any possible pair of adjacent training datasets. We consider $\mathcal{A_H}$ as the function $f$ in Definition \ref{def: sensitivity} to obtain the sensitivity of the machine learning algorithm that trains neural networks.

We have so far defined differentially private machine learning algorithms and the way of computing sensitivity.
We know from Remark \ref{remark:1-norm} that the ALM $M_{AL}(x, f, s)$ satisfies differential privacy if the output space of $f$ is a real coordinate space $\mathbb{R}^n$.
The machine learning algorithm in Definition \ref{prop:DP} is $\mathcal{A_H}: \mathcal{X} \mapsto \mathbb{R}^n$, whose input and output spaces are identical with $f$ in Remark \ref{remark:1-norm}.
Therefore, we deploy the ALM in the machine learning algorithm $\mathcal{A_H}$ and claim that $M_{AL}(x, \mathcal{A_H}, s)$ is an $\epsilon$-differentially private machine learning algorithm.
We can compute the privacy budget $\epsilon$ of the algorithm as the following:

\begin{proposition}
\label{prop:epsilon}
    The additive logistic mechanism $M_{AL} (x, \mathcal{A_H}, s)$ with machine learning algorithm $\mathcal{A_H}$ is $\epsilon$-differentially private for $\epsilon = \frac{\Delta_{\mathcal{H}}}{s}$.
    $\Delta_{\mathcal{H}}$ is the sensitivity of $\mathcal{A_H}$ and computed in the 1-norm distance.
\end{proposition}

\section{Post-training Protection on Self-Supervised Learning}

We have introduced the ALM and shown that the mechanism satisfy differential privacy in Section \ref{sec:alm}. We then show that we can construct a differentially private machine learning algorithm by deploying the ALM in a given machine learning algorithm in Section \ref{sec:app}.
In this section, we propose a post-training protection algorithm for deploying the ALM in self-supervised neural networks.

Recall that self-supervised learning contains two stages: pretraining stage and fine-tuning stage.
In the pretraining stage, the self-supervision algorithm $\mathcal{A_H}$ with the set of hyper-parameters $\mathcal{H}$ trains a base encoder $NN_{\mathcal{A_H}}$ by creating and fitting to pseudo-labels. In the fine-tuning stage, we apply an algorithm $\mathcal{A_S}$ with the set of hyper-parameters $\mathcal{S}$ to train a projection head $MP_{\mathcal{A_S}}$ to solve actual task.

We propose a post-training protection algorithm to provide differential privacy to $\mathcal{A_S}$. We implement an Additive Logistic Query Handler (ALQH) that deploys the ALM in constructing a differentially private machine learning algorithm $M_{AL}(\mathcal{A_S})$.

\begin{algorithm}
\caption{\textsc{Additive Logistic Query Handler}}
\label{algo: mechanism}
\KwIn{pretraining algorithm $\mathcal{A_H}$,
		pretraining dataset $D$,
		fine-tuning algorithm $\mathcal{A_S}$,
		fine-tuning dataset $d$,
		ALM parameter $s$,
		query point $x$}
\KwOut{protected output $\tilde y$}
$\theta \leftarrow \mathcal{A_\mathcal{H}}(D)$ \Comment*[r]{Train base encoder $NN_{\theta}$}

$\tilde r \leftarrow NN_{\theta} (d)$ \Comment*[r]{Obtain the outputs from the base encoder}

$\tilde \omega \sim \mathrm{M_{AL}(\tilde r, \mathcal{A_S}, s)}$ \Comment*[r]{Train the projection head $MP_{\omega}$ and add noise to its weights}

$\tilde y \sim MP_{\tilde \omega} ( NN_{\theta}(x) )$ \Comment*[r]{Output with respect to the query point $x$}
\end{algorithm}

The query handler applies the ALM to the weights $\omega$ of the projection head to preserve privacy in the fine-tuning stage. There is no need to re-train the base encoder $NN_\theta$ and the projection head $MP_\omega$ if we only adjust the parameter $s$ in the query handler without modifying the training algorithm and datasets.
We have shown that the mechanism $M_{AL}(x, \mathcal{A_S}, s)$ with hyper-parameter $s$ is $\epsilon$-differentially private in Proposition \ref{prop:epsilon}. Since we only apply the mechanism once to the weights $\omega$, there is no composition of privacy mechanisms.

We then need to compute the sensitivity $\Delta$ over the weights of the projection heads $\omega$ to determine the privacy budget $\epsilon$. Computing sensitivity is extremely time-consuming. From Definition \ref{def: sensitivity}, we have to compute the sensitivity by looping through every pair of adjacent datasets and train two networks for each pair of datasets.  Therefore, we use a Sensitivity Sampler (Algorithm \ref{algo:Sensitivity Sampler}) to estimate the sensitivity the self-supervised learning algorithm \citep{sensitivity_sampler}.
\begin{algorithm}[!ht]
\caption{\textsc{Sensitivity Sampler}}\label{algo:Sensitivity Sampler}
\KwIn{pretraining algorithm $\mathcal{A_H}$,
		pretraining dataset $D$,
		fine-tuning algorithm $\mathcal{A_S}$,
		fine-tuning dataset $d$,
		sample size $m$,
		random sampler $randint(\text{upper bound})$}
\KwOut{$\Delta$}

$\theta \leftarrow \mathcal{A_H}(D)$ \Comment*[r]{Train the base encoder $NN_{\theta}$}
\For{$k=1$ to $m$}{
$d_i, d_j \leftarrow d[randint(|d|)], d[randint(|d|)]$ \Comment*[r]{Select two random elements from the fine-tuning dataset $d$}
$D_1, D_2 \leftarrow d \setminus d_i, d \setminus d_j$ \;

$\omega_1, \omega_2 \leftarrow \mathcal{A_S}( NN_{\theta} (D_1) ), \mathcal{A_S}( NN_{\theta} (D_2) )$ \Comment*[r]{Fine-tune the models and obtain weights of the projection heads}

$\bar \Delta^{(i)} \leftarrow \| \omega_1 - \omega_2 \|_1$\;
}
$\Delta \leftarrow  \max_i \bar \Delta^{(i)}$\;
\end{algorithm}

\section{Empirical Analysis of Privacy-Performance Trade-off}
\label{sec:experiment}

The proposed post-training protection algorithm that deploys the ALM can lead to a loss in the neural network's performance while protecting its privacy. We define the term \textit{utility loss} to measure how the protection algorithm degrades the network's performance. We apply Algorithm \ref{algo: mechanism} to a set of queries from the validation dataset to obtain a set of protected outputs $\tilde y$. We feed the same set of queries to the unprotected neural networks to get a set of unprotected outputs $y$ and keep the ground truth labels $y_{GT}$ corresponding to these queries.
\begin{definition}
    \label{def:util_loss}
    Let $M$ be an evaluation metric that takes a set of predictions $y$ and its ground truth label set $y_{GT}$ as input and returns a numerical value to indicate the quality of the predictions. The \textbf{\textsc{Utility Loss}} caused by the protection algorithm is
    \begin{center}
        $\mathcal{L}_{util} = 1 - \frac{M(\tilde y, y_{GT})}{M(y, y_{GT})}$.
    \end{center}
\end{definition}

The base encoder $NN_\theta$ learns to generate \textit{representations}—projections of the input data onto the output space of $NN_\theta$.
A well-constructed base encoder maps the representations that consist of identical features into the same topological space \citep{Goodfellow-et-al-2016, bengio2014representation, yang2021}.
The projection head $MP_\omega$ divides the output space into a set of topological spaces and assigns each space a predicted label \citep{naitzat2020topology}.
Adding noise to the weights of the projection head modifies the boundary of the topological spaces and hence changes some of the label assignments.

Once the projection head is fully optimized, any changes in assignments leads to a loss of utility.
If the noise we add does not lead any individual to switch between spaces, then the noise will have a negligible impact on the outcomes. 
Conversely, large noise can push the representation beyond its original space and potentially lead to a loss in utility.
Therefore, it is desirable to have a mechanism that minimizes the amount of noise added in order to achieve a certain privacy level.
In this section, we empirically study relationship between the noise level, privacy budget, and utility loss.

\subsection{Privacy Budget vs. Utility Loss}
We performed the ALQH on Cifar 10 \citep{cifar}, Cifar 100 \citep{cifar}, and STL 10 \citep{deng2009imagenet} datasets to quantify the trade-off between privacy and performance of the self-supervised neural networks. We built the target image classifiers using Keras library \citep{chollet2015keras} and trained it following SimCLR \citep{chen2020simple} framework—one of the state of the arts self-supervised learning framework. We partitioned each dataset into pretraining and fine-tuning datasets, followed Algorithm \ref{algo: mechanism} to fine-tune the model, and added noise to the weights of fine-tuning layers.

\begin{figure}[t]
    \centering
    \resizebox{.33\linewidth}{!}{
\begin{tikzpicture}

\definecolor{darkgray176}{RGB}{176,176,176}
\definecolor{green}{RGB}{0,128,0}
\definecolor{lightgray204}{RGB}{204,204,204}
\definecolor{navy}{RGB}{0,0,128}
\definecolor{orange}{RGB}{255,165,0}
\definecolor{pink}{RGB}{255,192,203}

\begin{axis}[
legend cell align={left},
legend style={fill opacity=0.8, draw opacity=1, text opacity=1, draw=lightgray204},
log basis x={10},
tick align=outside,
tick pos=left,
x grid style={darkgray176},
xlabel={epsilon},
xmin=0.00973488200617172, xmax=16.939650595631,
xmode=log,
xtick style={color=black},
y grid style={darkgray176},
ylabel={Utility Loss},
ymin=-0.023748017360339, ymax=0.895199365093948,
ytick style={color=black}
]
\addplot [draw=white, fill=red, mark=*, only marks]
table{%
x  y
3.4984 0.0189793121301386
1.7492 0.024401916461318
0.8746 0.0349282374222313
0.4373 0.0677831351661986
0.21865 0.164752762473561
0.109325 0.347527920724482
0.0546625 0.661084519999259
0.02733125 0.772886767912139
0.013665625 0.80733651768838
};
\addlegendentry{Logistic}
\addplot [draw=white, fill=blue, mark=*, only marks]
table{%
x  y
3.4984 0.0215311019994806
1.7492 0.0248803696944193
0.8746 0.0248803696944193
0.4373 0.0558213594427519
0.21865 0.155661896275181
0.109325 0.332535875280888
0.0546625 0.628229669786907
0.02733125 0.814194581059938
0.013665625 0.853429029527844
};
\addlegendentry{Laplace}
\addplot [draw=white, fill=green, mark=*, only marks]
table{%
x  y
12.067175835298 0.0180223182057649
6.033587917649 0.0256778551250744
3.0167939588245 0.0326953276492762
1.50839697941225 0.039234442003606
0.754198489706125 0.0666666479582232
0.377099244853063 0.233014370646013
0.188549622426531 0.430941004882399
0.0942748112132656 0.633014344712764
0.0471374056066328 0.813397131277994
};
\addlegendentry{Gaussian}
\addplot [semithick, pink, forget plot]
table {%
0.013665625 0.80733651768838
0.02733125 0.772886767912139
0.0546625 0.661084519999259
0.109325 0.347527920724482
0.21865 0.164752762473561
0.4373 0.0677831351661986
0.8746 0.0349282374222313
1.7492 0.024401916461318
3.4984 0.0189793121301386
};
\addplot [semithick, navy, dash pattern=on 1pt off 3pt on 3pt off 3pt, forget plot]
table {%
0.013665625 0.853429029527844
0.02733125 0.814194581059938
0.0546625 0.628229669786907
0.109325 0.332535875280888
0.21865 0.155661896275181
0.4373 0.0558213594427519
0.8746 0.0248803696944193
1.7492 0.0248803696944193
3.4984 0.0215311019994806
};
\addplot [semithick, orange, dotted, forget plot]
table {%
0.0471374056066328 0.813397131277994
0.0942748112132656 0.633014344712764
0.188549622426531 0.430941004882399
0.377099244853063 0.233014370646013
0.754198489706125 0.0666666479582232
1.50839697941225 0.039234442003606
3.0167939588245 0.0326953276492762
6.033587917649 0.0256778551250744
12.067175835298 0.0180223182057649
};
\end{axis}

\end{tikzpicture}}
    \resizebox{.33\linewidth}{!}{
\begin{tikzpicture}

\definecolor{darkgray176}{RGB}{176,176,176}
\definecolor{green}{RGB}{0,128,0}
\definecolor{lightgray204}{RGB}{204,204,204}
\definecolor{navy}{RGB}{0,0,128}
\definecolor{orange}{RGB}{255,165,0}
\definecolor{pink}{RGB}{255,192,203}

\begin{axis}[
legend cell align={left},
legend style={fill opacity=0.8, draw opacity=1, text opacity=1, draw=lightgray204},
log basis x={10},
tick align=outside,
tick pos=left,
x grid style={darkgray176},
xlabel={epsilon},
xmin=0.0115420814098692, xmax=20.0578793068043,
xmode=log,
xtick style={color=black},
y grid style={darkgray176},
ylabel={Utility Loss},
ymin=-0.0462267151220137, ymax=1.00347583295023,
ytick style={color=black}
]
\addplot [draw=white, fill=red, mark=*, only marks]
table{%
x  y
4.1476 0.00148703706308817
2.0738 0.0211895850511289
1.0369 0.0791821346850201
0.51845 0.201115234633804
0.259225 0.453903336950394
0.1296125 0.697769515244048
0.06480625 0.864312266383916
0.032403125 0.928252785345436
0.0162015625 0.944237919461993
};
\addlegendentry{Logistic}
\addplot [draw=white, fill=blue, mark=*, only marks]
table{%
x  y
4.1476 0.00817848935889498
2.0738 0.0312267634948391
1.0369 0.071747230774408
0.51845 0.156877343096255
0.259225 0.423791817129766
0.1296125 0.683643113060068
0.06480625 0.868029735456169
0.032403125 0.934200744480449
0.0162015625 0.955762080765125
};
\addlegendentry{Laplace}
\addplot [draw=white, fill=green, mark=*, only marks]
table{%
x  y
14.289342516715 0.0133829276357885
7.14467125835752 0.0185873659126822
3.57233562917876 0.0412639419385492
1.78616781458938 0.113011149668782
0.89308390729469 0.277323429026125
0.446541953647345 0.579925657869715
0.223270976823672 0.829368035700241
0.111635488411836 0.920074346728041
0.0558177442059181 0.94832713877069
};
\addlegendentry{Gaussian}
\addplot [semithick, pink, forget plot]
table {%
0.0162015625 0.944237919461993
0.032403125 0.928252785345436
0.06480625 0.864312266383916
0.1296125 0.697769515244048
0.259225 0.453903336950394
0.51845 0.201115234633804
1.0369 0.0791821346850201
2.0738 0.0211895850511289
4.1476 0.00148703706308817
};
\addplot [semithick, navy, dash pattern=on 1pt off 3pt on 3pt off 3pt, forget plot]
table {%
0.0162015625 0.955762080765125
0.032403125 0.934200744480449
0.06480625 0.868029735456169
0.1296125 0.683643113060068
0.259225 0.423791817129766
0.51845 0.156877343096255
1.0369 0.071747230774408
2.0738 0.0312267634948391
4.1476 0.00817848935889498
};
\addplot [semithick, orange, dotted, forget plot]
table {%
0.0558177442059181 0.94832713877069
0.111635488411836 0.920074346728041
0.223270976823672 0.829368035700241
0.446541953647345 0.579925657869715
0.89308390729469 0.277323429026125
1.78616781458938 0.113011149668782
3.57233562917876 0.0412639419385492
7.14467125835752 0.0185873659126822
14.289342516715 0.0133829276357885
};
\end{axis}

\end{tikzpicture}}
    \resizebox{.33\linewidth}{!}{
\begin{tikzpicture}

\definecolor{darkgray176}{RGB}{176,176,176}
\definecolor{green}{RGB}{0,128,0}
\definecolor{lightgray204}{RGB}{204,204,204}
\definecolor{navy}{RGB}{0,0,128}
\definecolor{orange}{RGB}{255,165,0}
\definecolor{pink}{RGB}{255,192,203}

\begin{axis}[
legend cell align={left},
legend style={fill opacity=0.8, draw opacity=1, text opacity=1, draw=lightgray204},
log basis x={10},
tick align=outside,
tick pos=left,
x grid style={darkgray176},
xlabel={epsilon},
xmin=0.0073565969549732, xmax=13.3089343146826,
xmode=log,
xtick style={color=black},
y grid style={darkgray176},
ylabel={Utility Loss},
ymin=-0.0279912674229438, ymax=0.84980487940029,
ytick style={color=black}
]
\addplot [draw=white, fill=red, mark=*, only marks]
table{%
x  y
2.6484 0.0119085574326577
1.3242 0.0119086380053993
0.6621 0.0288058776045136
0.33105 0.0656179632832085
0.165525 0.186715475032672
0.0827625 0.420663013234183
0.04138125 0.653001282619517
0.020690625 0.738694875617099
0.0103453125 0.785162544457675
};
\addlegendentry{Logistic}
\addplot [draw=white, fill=blue, mark=*, only marks]
table{%
x  y
2.6484 0.0119085574326577
1.3242 0.0181445039504203
0.6621 0.0465079314615858
0.33105 0.0788944326952641
0.165525 0.172232068156674
0.0827625 0.38626486195373
0.04138125 0.571330857345979
0.020690625 0.750965554536943
0.0103453125 0.809905054544689
};
\addlegendentry{Laplace}
\addplot [draw=white, fill=green, mark=*, only marks]
table{%
x  y
9.46404138621551 0.0137189839635971
4.73202069310775 0.0261908846727167
2.36601034655388 0.0398696031377059
1.18300517327694 0.0630029289140015
0.591502586638469 0.159156777369926
0.295751293319235 0.28065658498125
0.147875646659617 0.502132301127983
0.0739378233298087 0.616189234349595
0.0369689116649043 0.718579005343299
};
\addlegendentry{Gaussian}
\addplot [semithick, pink, forget plot]
table {%
0.0103453125 0.785162544457675
0.020690625 0.738694875617099
0.04138125 0.653001282619517
0.0827625 0.420663013234183
0.165525 0.186715475032672
0.33105 0.0656179632832085
0.6621 0.0288058776045136
1.3242 0.0119086380053993
2.6484 0.0119085574326577
};
\addplot [semithick, navy, dash pattern=on 1pt off 3pt on 3pt off 3pt, forget plot]
table {%
0.0103453125 0.809905054544689
0.020690625 0.750965554536943
0.04138125 0.571330857345979
0.0827625 0.38626486195373
0.165525 0.172232068156674
0.33105 0.0788944326952641
0.6621 0.0465079314615858
1.3242 0.0181445039504203
2.6484 0.0119085574326577
};
\addplot [semithick, orange, dotted, forget plot]
table {%
0.0369689116649043 0.718579005343299
0.0739378233298087 0.616189234349595
0.147875646659617 0.502132301127983
0.295751293319235 0.28065658498125
0.591502586638469 0.159156777369926
1.18300517327694 0.0630029289140015
2.36601034655388 0.0398696031377059
4.73202069310775 0.0261908846727167
9.46404138621551 0.0137189839635971
};
\end{axis}

\end{tikzpicture}}
    \caption{Comparison between the ALM, the Gaussian mechanism, and the Laplace mechanism, in terms of the utility loss at each privacy budget. The plots from left to right are the results on Cifar 10, Cifar 100, and STL 10, respectively.}
    \label{fig:util_loss_eps}
\end{figure}
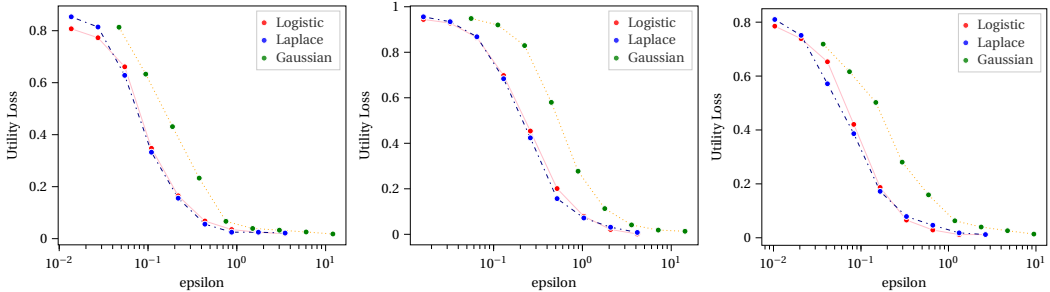

We computed \textit{empirical sensitivities} of the fine-tuned models for all three datasets, following Algorithm \ref{algo:Sensitivity Sampler}. We recorded the sensitivities and other experimental specifications in Table \ref{tab:util_spec}.
We took the sensitivity $\Delta$ and parameter $s$ of the ALM to compute the privacy budget $\epsilon$ based on Proposition \ref{prop:epsilon}. Then, we observed a negative correlation between the privacy budget and the model's utility in Figure \ref{fig:util_loss_eps}.

\begin{table}[!ht]
    \centering
    \caption{Specifications on the fine-tuned self-supervised models. $m$ is the sample size of the Sensitivity Sampler. |D| is the size of pretraining dataset and |d| is the size of fine-tuning dataset.}
    \begin{tabular}{||c|c|c|c|c|c||}
         \hline
         Dataset & |D| & |d| & m & 1-norm sensitivity & 2-norm sensitivity \\
         \hline
         Cifar 10 & 40000 & 10000 & 500 & 0.017492 & 0.013842 \\
         Cifar 100 & 40000 & 10000 & 500 & 0.020738 & 0.016391 \\
         STL 10 & 100000 & 5000 & 250 & 0.013242 & 0.010856\\
         \hline
    \end{tabular}
    \label{tab:util_spec}
\end{table}

We then repeated the experiments on two of the most well-known additive noise mechanisms: the Laplace mechanism and the Gaussian mechanism. We applied Algorithm \ref{algo: mechanism} but replaced the ALM $M_{AL}(\tilde r, \mathcal{A_S}, s)$ in the algorithm with the Gaussian $M_{Gauss}(\tilde r, \mathcal{A_S}, \sigma)$ and the Laplace mechanisms $M_{Lap}(\tilde r, \mathcal{A_S}, b)$, respectively. $\sigma$ and $b$ are the distribution parameters stated in Remark \ref{remark:anm}.

The Gaussian mechanism $M_{Gauss}$ is ($\epsilon, \delta$)-differentially private; we chose $\delta=1e-5$ for all the experiments and used the 2-norm sensitivity to compute $\epsilon$. The Laplace mechanism and the ALM are $\epsilon$-differentially private; hence, we chose $\delta=0$ and used the 1-norm sensitivity in the experiments.

We compare the privacy-utility trade-off of the three additive noise mechanisms in Figure \ref{fig:util_loss_eps}.
We observe that the utility loss caused by the ALM was approximately equal to the utility loss by the Laplace mechanism at each privacy budget tested, whereas the Gaussian mechanism led to a significantly higher utility loss.

The Gaussian mechanism is $(\epsilon, \delta)$-differentially privacy, while the Laplace mechanism and the ALM are $\epsilon$-differentially private. 
To maintain the same privacy budget $\epsilon$, the Gaussian mechanism has to draw noise from a distribution with a higher variance compared with the distributions of the other two mechanisms.
Sampling noise from a distribution with a larger variance amplifies the changes in label assignments, which leads to a larger utility loss.

\subsection{Privacy Protection against Membership Inference Attack}
The post-training protection algorithm with the ALM could effectively protect self-supervised models against MIAs.
We first performed MIAs on the queries from the fine-tuning datasets with the unprotected outputs to acquire benchmark attack accuracies. We got benchmark MIA accuracies of \textbf{62\%} on Cifar 10, \textbf{61\%} on STL 10, and \textbf{71\%} on Cifar 100.
Then, we performed MIAs on the same set of queries with the protected outputs obtained from Algorithm \ref{algo: mechanism}. We state the implementation and experimental setup of MIA in the Appendix.
Our experiments shows that the ALM is able to effectively reduce the MIA accuracy to approximately 50\%, which is equivalent to random guessing. 

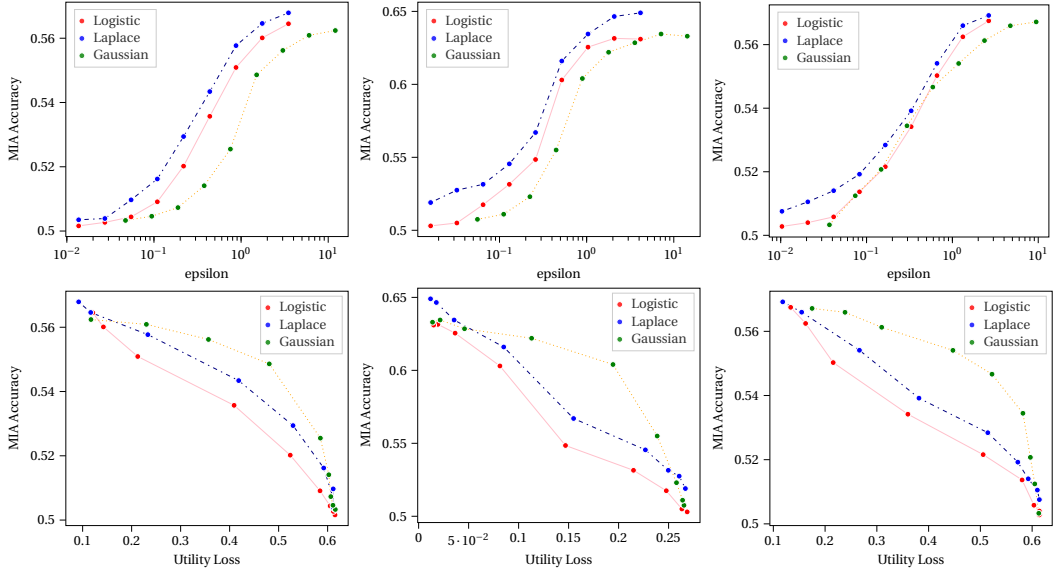
\begin{figure}[t]
    \centering
    \resizebox{.33\linewidth}{!}{
\begin{tikzpicture}

\definecolor{darkgray176}{RGB}{176,176,176}
\definecolor{green}{RGB}{0,128,0}
\definecolor{lightgray204}{RGB}{204,204,204}
\definecolor{navy}{RGB}{0,0,128}
\definecolor{orange}{RGB}{255,165,0}
\definecolor{pink}{RGB}{255,192,203}

\begin{axis}[
legend cell align={left},
legend style={
  fill opacity=0.8,
  draw opacity=1,
  text opacity=1,
  at={(0.03,0.97)},
  anchor=north west,
  draw=lightgray204
},
log basis x={10},
tick align=outside,
tick pos=left,
x grid style={darkgray176},
xlabel={epsilon},
xmin=0.00973488200617172, xmax=16.939650595631,
xmode=log,
xtick style={color=black},
y grid style={darkgray176},
ylabel={MIA Accuracy},
ymin=0.498285, ymax=0.571215,
ytick style={color=black}
]
\addplot [draw=white, fill=red, mark=*, only marks]
table{%
x  y
3.4984 0.5645
1.7492 0.5601
0.8746 0.5509
0.4373 0.5357
0.21865 0.5202
0.109325 0.5091
0.0546625 0.5044
0.02733125 0.5027
0.013665625 0.5016
};
\addlegendentry{Logistic}
\addplot [draw=white, fill=blue, mark=*, only marks]
table{%
x  y
3.4984 0.5679
1.7492 0.5646
0.8746 0.5577
0.4373 0.5434
0.21865 0.5294
0.109325 0.5162
0.0546625 0.5097
0.02733125 0.5039
0.013665625 0.5035
};
\addlegendentry{Laplace}
\addplot [draw=white, fill=green, mark=*, only marks]
table{%
x  y
12.067175835298 0.5624
6.033587917649 0.5609
3.0167939588245 0.5562
1.50839697941225 0.5486
0.754198489706125 0.5255
0.377099244853063 0.5141
0.188549622426531 0.5073
0.0942748112132656 0.5046
0.0471374056066328 0.5033
};
\addlegendentry{Gaussian}
\addplot [semithick, pink, forget plot]
table {%
0.013665625 0.5016
0.02733125 0.5027
0.0546625 0.5044
0.109325 0.5091
0.21865 0.5202
0.4373 0.5357
0.8746 0.5509
1.7492 0.5601
3.4984 0.5645
};
\addplot [semithick, navy, dash pattern=on 1pt off 3pt on 3pt off 3pt, forget plot]
table {%
0.013665625 0.5035
0.02733125 0.5039
0.0546625 0.5097
0.109325 0.5162
0.21865 0.5294
0.4373 0.5434
0.8746 0.5577
1.7492 0.5646
3.4984 0.5679
};
\addplot [semithick, orange, dotted, forget plot]
table {%
0.0471374056066328 0.5033
0.0942748112132656 0.5046
0.188549622426531 0.5073
0.377099244853063 0.5141
0.754198489706125 0.5255
1.50839697941225 0.5486
3.0167939588245 0.5562
6.033587917649 0.5609
12.067175835298 0.5624
};
\end{axis}

\end{tikzpicture}}
    \resizebox{.33\linewidth}{!}{
\begin{tikzpicture}

\definecolor{darkgray176}{RGB}{176,176,176}
\definecolor{green}{RGB}{0,128,0}
\definecolor{lightgray204}{RGB}{204,204,204}
\definecolor{navy}{RGB}{0,0,128}
\definecolor{orange}{RGB}{255,165,0}
\definecolor{pink}{RGB}{255,192,203}

\begin{axis}[
legend cell align={left},
legend style={
  fill opacity=0.8,
  draw opacity=1,
  text opacity=1,
  at={(0.03,0.97)},
  anchor=north west,
  draw=lightgray204
},
log basis x={10},
tick align=outside,
tick pos=left,
x grid style={darkgray176},
xlabel={epsilon},
xmin=0.0115420814098692, xmax=20.0578793068043,
xmode=log,
xtick style={color=black},
y grid style={darkgray176},
ylabel={MIA Accuracy},
ymin=0.495700002074242, ymax=0.656299998402596,
ytick style={color=black}
]
\addplot [draw=white, fill=red, mark=*, only marks]
table{%
x  y
4.1476 0.630999982357025
2.0738 0.631500005722046
1.0369 0.625500013828278
0.51845 0.602999985218048
0.259225 0.548500001430511
0.1296125 0.531500010490418
0.06480625 0.517499983310699
0.032403125 0.504999985694885
0.0162015625 0.503000001907349
};
\addlegendentry{Logistic}
\addplot [draw=white, fill=blue, mark=*, only marks]
table{%
x  y
4.1476 0.648999998569489
2.0738 0.646500012397766
1.0369 0.634500026702881
0.51845 0.61599999666214
0.259225 0.566999971866608
0.1296125 0.54550002861023
0.06480625 0.531500002384186
0.032403125 0.52749998998642
0.0162015625 0.51900000333786
};
\addlegendentry{Laplace}
\addplot [draw=white, fill=green, mark=*, only marks]
table{%
x  y
14.289342516715 0.633000016212463
7.14467125835752 0.634500026702881
3.57233562917876 0.628500025272369
1.78616781458938 0.621999979019165
0.89308390729469 0.603999972343445
0.446541953647345 0.555000007152557
0.223270976823672 0.523000020980835
0.111635488411836 0.510999987125397
0.0558177442059181 0.507499992847443
};
\addlegendentry{Gaussian}
\addplot [semithick, pink, forget plot]
table {%
0.0162015625 0.503000001907349
0.032403125 0.504999985694885
0.06480625 0.517499983310699
0.1296125 0.531500010490418
0.259225 0.548500001430511
0.51845 0.602999985218048
1.0369 0.625500013828278
2.0738 0.631500005722046
4.1476 0.630999982357025
};
\addplot [semithick, navy, dash pattern=on 1pt off 3pt on 3pt off 3pt, forget plot]
table {%
0.0162015625 0.51900000333786
0.032403125 0.52749998998642
0.06480625 0.531500002384186
0.1296125 0.54550002861023
0.259225 0.566999971866608
0.51845 0.61599999666214
1.0369 0.634500026702881
2.0738 0.646500012397766
4.1476 0.648999998569489
};
\addplot [semithick, orange, dotted, forget plot]
table {%
0.0558177442059181 0.507499992847443
0.111635488411836 0.510999987125397
0.223270976823672 0.523000020980835
0.446541953647345 0.555000007152557
0.89308390729469 0.603999972343445
1.78616781458938 0.621999979019165
3.57233562917876 0.628500025272369
7.14467125835752 0.634500026702881
14.289342516715 0.633000016212463
};
\end{axis}

\end{tikzpicture}}
    \resizebox{.33\linewidth}{!}{
\begin{tikzpicture}

\definecolor{darkgray176}{RGB}{176,176,176}
\definecolor{green}{RGB}{0,128,0}
\definecolor{lightgray204}{RGB}{204,204,204}
\definecolor{navy}{RGB}{0,0,128}
\definecolor{orange}{RGB}{255,165,0}
\definecolor{pink}{RGB}{255,192,203}

\begin{axis}[
legend cell align={left},
legend style={
  fill opacity=0.8,
  draw opacity=1,
  text opacity=1,
  at={(0.03,0.97)},
  anchor=north west,
  draw=lightgray204
},
log basis x={10},
tick align=outside,
tick pos=left,
x grid style={darkgray176},
xlabel={epsilon},
xmin=0.0073565969549732, xmax=13.3089343146826,
xmode=log,
xtick style={color=black},
y grid style={darkgray176},
ylabel={MIA Accuracy},
ymin=0.499465969, ymax=0.572504671,
ytick style={color=black}
]
\addplot [draw=white, fill=red, mark=*, only marks]
table{%
x  y
2.6484 0.56748546
1.3242 0.56245745
0.6621 0.55026287
0.33105 0.53417408
0.165525 0.52158091
0.0827625 0.5136828
0.04138125 0.50581111
0.020690625 0.50399493
0.0103453125 0.50278591
};
\addlegendentry{Logistic}
\addplot [draw=white, fill=blue, mark=*, only marks]
table{%
x  y
2.6484 0.56918473
1.3242 0.56597119
0.6621 0.55410363
0.33105 0.53917905
0.165525 0.52841136
0.0827625 0.51923413
0.04138125 0.51404148
0.020690625 0.51053137
0.0103453125 0.50755216
};
\addlegendentry{Laplace}
\addplot [draw=white, fill=green, mark=*, only marks]
table{%
x  y
9.46404138621551 0.56714533
4.73202069310775 0.56593257
2.36601034655388 0.56126741
1.18300517327694 0.55407091
0.591502586638469 0.5466455
0.295751293319235 0.53446105
0.147875646659617 0.52073309
0.0739378233298087 0.51244786
0.0369689116649043 0.50330254
};
\addlegendentry{Gaussian}
\addplot [semithick, pink, forget plot]
table {%
0.0103453125 0.50278591
0.020690625 0.50399493
0.04138125 0.50581111
0.0827625 0.5136828
0.165525 0.52158091
0.33105 0.53417408
0.6621 0.55026287
1.3242 0.56245745
2.6484 0.56748546
};
\addplot [semithick, navy, dash pattern=on 1pt off 3pt on 3pt off 3pt, forget plot]
table {%
0.0103453125 0.50755216
0.020690625 0.51053137
0.04138125 0.51404148
0.0827625 0.51923413
0.165525 0.52841136
0.33105 0.53917905
0.6621 0.55410363
1.3242 0.56597119
2.6484 0.56918473
};
\addplot [semithick, orange, dotted, forget plot]
table {%
0.0369689116649043 0.50330254
0.0739378233298087 0.51244786
0.147875646659617 0.52073309
0.295751293319235 0.53446105
0.591502586638469 0.5466455
1.18300517327694 0.55407091
2.36601034655388 0.56126741
4.73202069310775 0.56593257
9.46404138621551 0.56714533
};
\end{axis}

\end{tikzpicture}}
    
    \resizebox{.33\linewidth}{!}{
\begin{tikzpicture}

\definecolor{darkgray176}{RGB}{176,176,176}
\definecolor{green}{RGB}{0,128,0}
\definecolor{lightgray204}{RGB}{204,204,204}
\definecolor{navy}{RGB}{0,0,128}
\definecolor{orange}{RGB}{255,165,0}
\definecolor{pink}{RGB}{255,192,203}

\begin{axis}[
legend cell align={left},
legend style={fill opacity=0.8, draw opacity=1, text opacity=1, draw=lightgray204},
tick align=outside,
tick pos=left,
x grid style={darkgray176},
xlabel={Utility Loss},
xmin=0.0657099980860949, xmax=0.641890006884932,
xtick style={color=black},
y grid style={darkgray176},
ylabel={MIA Accuracy},
ymin=0.498285, ymax=0.571215,
ytick style={color=black}
]
\addplot [draw=white, fill=red, mark=*, only marks]
table{%
x  y
0.615099971294403 0.5016
0.611699998378754 0.5027
0.605099995136261 0.5044
0.584499974250793 0.5091
0.523700017929077 0.5202
0.40909999370575 0.5357
0.212500005960464 0.5509
0.142399996519089 0.5601
0.120800003409386 0.5645
};
\addlegendentry{Logistic}
\addplot [draw=white, fill=blue, mark=*, only marks]
table{%
x  y
0.613499999046326 0.5035
0.611400008201599 0.5039
0.611400008201599 0.5097
0.592000007629395 0.5162
0.529399991035461 0.5294
0.418500006198883 0.5434
0.23309999704361 0.5577
0.116499997675419 0.5646
0.091899998486042 0.5679
};
\addlegendentry{Laplace}
\addplot [draw=white, fill=green, mark=*, only marks]
table{%
x  y
0.615700006484985 0.5033
0.610899984836578 0.5046
0.606500029563904 0.5073
0.602400004863739 0.5141
0.585200011730194 0.5255
0.48089998960495 0.5486
0.356799989938736 0.5562
0.230100005865097 0.5609
0.116999998688698 0.5624
};
\addlegendentry{Gaussian}
\path [draw=navy, fill=navy, opacity=0.2]
(axis cs:0.611400008201599,0.5097)
--(axis cs:0.611400008201599,0.5039)
--(axis cs:0.611400008201599,0.5097)
--(axis cs:0.611400008201599,0.5097)
--cycle;

\addplot [semithick, pink, forget plot]
table {%
0.120800003409386 0.5645
0.142399996519089 0.5601
0.212500005960464 0.5509
0.40909999370575 0.5357
0.523700017929077 0.5202
0.584499974250793 0.5091
0.605099995136261 0.5044
0.611699998378754 0.5027
0.615099971294403 0.5016
};
\addplot [semithick, navy, dash pattern=on 1pt off 3pt on 3pt off 3pt, forget plot]
table {%
0.091899998486042 0.5679
0.116499997675419 0.5646
0.23309999704361 0.5577
0.418500006198883 0.5434
0.529399991035461 0.5294
0.592000007629395 0.5162
0.611400008201599 0.5068
0.613499999046326 0.5035
};
\addplot [semithick, orange, dotted, forget plot]
table {%
0.116999998688698 0.5624
0.230100005865097 0.5609
0.356799989938736 0.5562
0.48089998960495 0.5486
0.585200011730194 0.5255
0.602400004863739 0.5141
0.606500029563904 0.5073
0.610899984836578 0.5046
0.615700006484985 0.5033
};
\end{axis}

\end{tikzpicture}}
    \resizebox{.33\linewidth}{!}{
\begin{tikzpicture}

\definecolor{darkgray176}{RGB}{176,176,176}
\definecolor{green}{RGB}{0,128,0}
\definecolor{lightgray204}{RGB}{204,204,204}
\definecolor{navy}{RGB}{0,0,128}
\definecolor{orange}{RGB}{255,165,0}
\definecolor{pink}{RGB}{255,192,203}

\begin{axis}[
legend cell align={left},
legend style={fill opacity=0.8, draw opacity=1, text opacity=1, draw=lightgray204},
tick align=outside,
tick pos=left,
x grid style={darkgray176},
xlabel={Utility Loss},
xmin=-0.000934999063611032, xmax=0.281434986367822,
xtick style={color=black},
y grid style={darkgray176},
ylabel={MIA Accuracy},
ymin=0.495700002074242, ymax=0.656299998402596,
ytick style={color=black}
]
\addplot [draw=white, fill=red, mark=*, only marks]
table{%
x  y
0.268599987030029 0.503000001907349
0.263300001621246 0.504999985694885
0.24770000576973 0.517499983310699
0.214900001883507 0.531500010490418
0.146900002360344 0.548500001430511
0.0813000003993511 0.602999985218048
0.0365000003427267 0.625500013828278
0.0193000007420778 0.631500005722046
0.0149999996647239 0.630999982357025
};
\addlegendentry{Logistic}
\addplot [draw=white, fill=blue, mark=*, only marks]
table{%
x  y
0.266799986362457 0.51900000333786
0.260600000619888 0.52749998998642
0.249699994921684 0.531500002384186
0.226799994707108 0.54550002861023
0.155000001192093 0.566999971866608
0.0851000025868416 0.61599999666214
0.0355000011622906 0.634500026702881
0.0176999997347593 0.646500012397766
0.0119000002741814 0.648999998569489
};
\addlegendentry{Laplace}
\addplot [draw=white, fill=green, mark=*, only marks]
table{%
x  y
0.265399992465973 0.507499992847443
0.263999998569489 0.510999987125397
0.25789999961853 0.523000020980835
0.238600000739098 0.555000007152557
0.194399997591972 0.603999972343445
0.112999998033047 0.621999979019165
0.0458999983966351 0.628500025272369
0.0215000007301569 0.634500026702881
0.0138999996706843 0.633000016212463
};
\addlegendentry{Gaussian}
\addplot [semithick, pink, forget plot]
table {%
0.0149999996647239 0.630999982357025
0.0193000007420778 0.631500005722046
0.0365000003427267 0.625500013828278
0.0813000003993511 0.602999985218048
0.146900002360344 0.548500001430511
0.214900001883507 0.531500010490418
0.24770000576973 0.517499983310699
0.263300001621246 0.504999985694885
0.268599987030029 0.503000001907349
};
\addplot [semithick, navy, dash pattern=on 1pt off 3pt on 3pt off 3pt, forget plot]
table {%
0.0119000002741814 0.648999998569489
0.0176999997347593 0.646500012397766
0.0355000011622906 0.634500026702881
0.0851000025868416 0.61599999666214
0.155000001192093 0.566999971866608
0.226799994707108 0.54550002861023
0.249699994921684 0.531500002384186
0.260600000619888 0.52749998998642
0.266799986362457 0.51900000333786
};
\addplot [semithick, orange, dotted, forget plot]
table {%
0.0138999996706843 0.633000016212463
0.0215000007301569 0.634500026702881
0.0458999983966351 0.628500025272369
0.112999998033047 0.621999979019165
0.194399997591972 0.603999972343445
0.238600000739098 0.555000007152557
0.25789999961853 0.523000020980835
0.263999998569489 0.510999987125397
0.265399992465973 0.507499992847443
};
\end{axis}

\end{tikzpicture}}
    \resizebox{.33\linewidth}{!}{
\begin{tikzpicture}

\definecolor{darkgray176}{RGB}{176,176,176}
\definecolor{green}{RGB}{0,128,0}
\definecolor{lightgray204}{RGB}{204,204,204}
\definecolor{navy}{RGB}{0,0,128}
\definecolor{orange}{RGB}{255,165,0}
\definecolor{pink}{RGB}{255,192,203}

\begin{axis}[
legend cell align={left},
legend style={fill opacity=0.8, draw opacity=1, text opacity=1, draw=lightgray204},
tick align=outside,
tick pos=left,
x grid style={darkgray176},
xlabel={Utility Loss},
xmin=0.0933312479406595, xmax=0.638793773576617,
xtick style={color=black},
y grid style={darkgray176},
ylabel={MIA Accuracy},
ymin=0.499465969, ymax=0.572504671,
ytick style={color=black}
]
\addplot [draw=white, fill=red, mark=*, only marks]
table{%
x  y
0.614000022411346 0.50278591
0.613999972343445 0.50399493
0.603500027656555 0.50581111
0.580624997615814 0.5136828
0.505375003814697 0.52158091
0.360000003576279 0.53417408
0.215625002980232 0.55026287
0.162375004291534 0.56245745
0.133499994874001 0.56748546
};
\addlegendentry{Logistic}
\addplot [draw=white, fill=blue, mark=*, only marks]
table{%
x  y
0.614000022411346 0.50755216
0.610125005245209 0.51053137
0.592499971389771 0.51404148
0.572374999523163 0.51923413
0.514374992847443 0.52841136
0.381375014781952 0.53917905
0.266375005245209 0.55410363
0.154750004410744 0.56597119
0.11812499910593 0.56918473
};
\addlegendentry{Laplace}
\addplot [draw=white, fill=green, mark=*, only marks]
table{%
x  y
0.612875023365021 0.50330254
0.605124984264374 0.51244786
0.596625028610229 0.52073309
0.582249979972839 0.53446105
0.522499978542328 0.5466455
0.446999998092651 0.55407091
0.309374988079071 0.56126741
0.238500009775162 0.56593257
0.174875006079674 0.56714533
};
\addlegendentry{Gaussian}
\addplot [semithick, pink, forget plot]
table {%
0.133499994874001 0.56748546
0.162375004291534 0.56245745
0.215625002980232 0.55026287
0.360000003576279 0.53417408
0.505375003814697 0.52158091
0.580624997615814 0.5136828
0.603500027656555 0.50581111
0.613999972343445 0.50399493
0.614000022411346 0.50278591
};
\addplot [semithick, navy, dash pattern=on 1pt off 3pt on 3pt off 3pt, forget plot]
table {%
0.11812499910593 0.56918473
0.154750004410744 0.56597119
0.266375005245209 0.55410363
0.381375014781952 0.53917905
0.514374992847443 0.52841136
0.572374999523163 0.51923413
0.592499971389771 0.51404148
0.610125005245209 0.51053137
0.614000022411346 0.50755216
};
\addplot [semithick, orange, dotted, forget plot]
table {%
0.174875006079674 0.56714533
0.238500009775162 0.56593257
0.309374988079071 0.56126741
0.446999998092651 0.55407091
0.522499978542328 0.5466455
0.582249979972839 0.53446105
0.596625028610229 0.52073309
0.605124984264374 0.51244786
0.612875023365021 0.50330254
};
\end{axis}

\end{tikzpicture}}
    \caption{The top row shows the MIA accuracy at each privacy budget $\epsilon$. The bottom row shows the trade-off between the MIA accuracy and the model's utility loss. The columns from the left to right present the results on Cifar 10, Cifar 100, and STL 10, respectively.}
    \label{fig:mia_eps}
\end{figure}

Then, we replaced the ALM in the Algorithm \ref{algo: mechanism} with the Gaussian mechanism and the Laplace mechanism, respectively, and repeated the MIA experiments. We compared the performance of the Gaussian mechanism and the Laplace mechanism with the ALM in protecting against MIAs. We present the results in Figure \ref{fig:mia_eps}.

We have two observations from Figure \ref{fig:mia_eps}: first, the Gaussian mechanism and the ALM achieve the MIA accuracies closer to 50\% at each privacy budget, while the Laplace mechanism maintains the highest MIA accuracy.


Stronger protection leads to a lower MIA accuracy, simultaneously, it also results in a bigger utility loss to the neural network.
We quantify this trade-off between the MIA accuracy and utility loss, named the protection-utility trade-off.
We have shown and justified that the Gaussian mechanism could lead to a significantly higher utility loss than the other two mechanisms at the same privacy budget $\epsilon$.

As for the second observation, the ALM is able to maintain the lowest utility loss at a fixed MIA accuracy compared to Gaussian and Laplace mechanisms.
The ALM provides slightly weaker protections against MIAs compared to the Gaussian mechanism. However, the utility loss caused by the ALM is significantly lower than the Gaussian mechanism. Hence we can observe that the ALM achieves a better protection-utility trade-off.
On the other hand, the ALM maintains approximately equal utility loss with the Laplace mechanism, while the ALM is able to reduce the MIA accuracy closer to 50\%.
Therefore, we conclude that the ALM outperforms the other two mechanisms in protecting against MIAs since it achieves the best protection-utility trade-off.

\section{Conclusion}
In this work, we introduce a privacy mechanism named additive logistic mechanism. We theoretically prove that the additive logistic mechanism satisfies differential privacy and compute the privacy budget in terms of the parameters of the logistic distribution.
We then propose a post-training protection algorithm for the fine-tuning stage of the self-supervised learning algorithm. The protection algorithm deploys the additive logistic mechanism to enforce differential privacy to the fine-tuning data.
We have verified the effectiveness of our proposed protection algorithm with the additive logistic mechanism by showing the decreasing MIA accuracies on the protected models.
Finally, we demonstrate how the additive logistic mechanism stands out from the existing additive noise mechanisms in protecting self-supervised learning algorithms. 
A limitation of the post-training protection with the additive logistic mechanism is that the utility loss can increase as more layers in the protected neural network.
As an extension to this work, we are interested in developing a moment accountant for this mechanism for adapting DP-SGD, which is less sensitive to the scale of the neural network.

\bibliographystyle{plainnat}
\bibliography{references}

\newpage

\appendix
\section{Appendix}
\subsection{Reproducibility}

We use SimCLR to train our self-supervised models in the experiments. The implementations follows the code from Keras library (\url{https://keras.io/examples/vision/semisupervised_simclr/}) with default hyper-parameters.

For the implementation of the MIA, we use an MLP with 5 ReLU-activated hidden layers. We train the classifier using 2,000 probability pairs with half labeled `in'. We use the Keras library to train the binary classifier with learning rate 0.001 and default parameters.

All the experiments are done using a single Nvidia K80 / T4 GPU.

\section{Membership Inference Attack}
First, the attacker construct a shadow model to mimic the \textsc{victim model}'s behavior without directly having access to its training dataset or its learning algorithm.

In the next step, the attacker splits its training dataset into two partitions from which it uses only one to train the shadow model. 

After that, The attacker collects the shadow model's outputs corresponding to the seen and the unseen data records and trains a binary classifier to predict if a data record is 'in' or 'out' of the training dataset of the shadow model.

In our later experiments, we use a 3-tuple format to populate the binary classifier's training dataset.
The first element of the tuple is the output probability vectors that the shadow model generates from either partitions of its training dataset.
The second element consists of the ground truth labels about the output in the first element to help the MIA distinguish between the members and non-members.
The third element is a label that indicates which partition of datasets this record belongs to. 
Once the classifier is well-trained, we apply it to the \textsc{victim model} and infer the its training data.

\section{Proofs}
\subsection{Proof on Theorem \ref{thm:epsilon}}
\begin{proof}
    Let $f: \mathcal{X} \longrightarrow \mathbb{R}$ be a function, take arbitrary $x, y$ from $\mathcal{X}$, then we can compute the privacy budget of our mechanism $M_{AL}$. Refer to definition \ref{def:DP}, take any $z \in \mathcal{Y}$, we can show that our mechanism is differentially private by showing there exists an $\epsilon$ such that:
    \begin{center}
        $\mathrm{P}[M_{AL} (x, f, s) = z] \le \exp (\epsilon) \mathrm{P}[M_{AL} (y, f, s) = z]$.
    \end{center}
    This is equivalent to showing there exists an $\epsilon$ such that:
    \begin{equation}
        \log \left(\frac{\mathrm{P}[M_{AL} (x, f, s) = z]}{\mathrm{P}[M_{AL} (y, f, s)=z]} \right) \le \epsilon
    \end{equation}
    Then we can start the proof:
    \begin{equation}
        \label{proof:dp}
        \begin{split}
        \log \left(\frac{\mathrm{P}[M_{AL} (x, f, s) = z]}{\mathrm{P}[M_{AL} (y, f, s)=z]} \right) & = \log(\mathrm{P}[M_{AL} (x, f, s) = z]) - \log(\mathrm{P}[M_{AL} (y, f, s)=z]) \\
        & = \log(\mathrm{P}[f(x)+n=z])-\log(\mathrm{P}[f(y)+n=z])\\
        & = \log(\mathrm{P}[n=z-f(x)]) - \log(\mathrm{P}[n=z-f(y)]) \\
        & = \log \left(\frac{\exp(\frac{-(z-f(x))}{s})}{s(1 + \exp(\frac{-(z-f(x))}{s}) )^2} \right) - \log\left(\frac{\exp(\frac{-(z-f(x))}{s})}{s(1 + \exp(\frac{-(z-f(x))}{s}) )^2} \right)\\
        & = \frac{-(z-f(x))}{s} - \log(s) - 2 \log(1 + \exp(\frac{-(z-f(x))}{s}))\\
        & - \frac{-(z-f(y))}{s} +\log(s) + 2 \log(1 + \exp(\frac{-(z-f(x))}{s})) \\
        & = \frac{f(x) - f(y)}{s} + 2 \log \left( \frac{1 + \exp(\frac{f(y)-z}{s})}{1 + \exp(\frac{f(x)-z}{s})} \right)
        \end{split}
    \end{equation}
    
    Let $\gamma = |f(x) - f(y)|$, then we have two cases: $f(x) \ge f(y)$ or $f(x) < f(y)$. 
    
    In the first case, $\gamma = |f(x) - f(y)| = f(x) - f(y)$, then $f(x) = f(y) + \gamma$. We substitute $f(x)$ in Equation \ref{proof:dp} with $f(y) + \gamma$ and get
    \begin{equation}
        \label{eq:case1}
        \frac{\gamma}{s} + 2 \log \left( \frac{1 + \exp(\frac{f(y)-z}{s})}{1 + \exp(\frac{f(y)+\gamma-z}{s})} \right) \le \frac{\gamma}{s} + 2 \log \left( \frac{1 + \exp(\frac{f(y)-z}{s})}{1 + \exp(\frac{f(y)-z}{s})} \right) = \frac{\gamma}{s}.
    \end{equation}
    
    In the second case, $\gamma = |f(x) - f(y)| = f(y) - f(x)$. We substitute $f(y)$ Equation \ref{proof:dp} with $f(x) + \gamma$ and get
    \begin{equation}
        \label{eq:case2}
        \begin{split}
            \frac{-\gamma}{s} + 2 \log \left( \frac{1 + \exp(\frac{f(x)+\gamma -z}{s})}{1 + \exp(\frac{f(x)-z}{s})} \right) & = \frac{-\gamma}{s} + 2 \log \left( \frac{\exp(\gamma) (\frac{1}{\exp(\gamma)} +  \exp(\frac{f(x)-z}{s})) }{1 + \exp(\frac{f(x)-z}{s})} \right) \\
            & \le \frac{-\gamma}{s} + 2 \log \left( \frac{\exp(\frac{\gamma}{s}) (1 +  \exp(\frac{f(x)-z}{s})) }{1 + \exp(\frac{f(x)-z}{s})} \right) \\
            & = \frac{-\gamma}{s} + 2 \log (\exp(\gamma/s)) = \frac{-\gamma}{s} + 2 \frac{\gamma}{s} = \frac{\gamma}{s}.
        \end{split}
    \end{equation}
    
    In both cases, we get a bound of $\frac{\gamma}{s}$. From \textbf{Definition \ref{def: sensitivity}}, we know $\gamma = |f(x) - f(y)| \le \Delta$, where $\Delta$ is the sensitivity of $f$. Therefore, we can get $\epsilon = \frac{\Delta}{s}$. Hence we have proved that our additive exponential mechanism is $\frac{\lambda \Delta}{c}$-differentially private.
\end{proof}

\subsection{Proof on Remark \ref{remark:1-norm}}
\begin{proof}
    Given a function $f: \mathcal{X} \longrightarrow \mathbb{R}^n$, similar to the previous proof, we want to show that 
    \begin{center}
        $\mathrm{P}[M_{AL} (x, f, s) = \Vec{z}] \le \exp (\epsilon) \mathrm{P}[M_{AL} (y, f, s) = \Vec{z}]$.
    \end{center}
    Denote $\Delta_i = |f(x)_i - f(y)_i|$ for $i \in \mathbb{Z}^+$ and $i <= n$. We move the probabilities to one side and continue the proof:
    \begin{equation} 
    \label{proof:dp2}
    \begin{split}
    \frac{\mathrm{P}[M_{AL} (x, f, s) = \Vec{z}]}{\mathrm{P}[M_{AL} (y, f, s) = \Vec{z}]} 
     & = \frac{ \prod_{i=1}^n  \mathrm{P}[f(x)_i + Log(s)_i = \Vec{z}_i]}{\prod_{i=1}^n \mathrm{P}[f(y)_i + Log(s)_i = \Vec{z}_i]} \\
     & = \prod_{i=1}^n \frac{\mathrm{P}[Log(s)_i = \Vec{z}_i - f(x)_i]}{\mathrm{P}[Log(s)_i = \Vec{z}_i - f(y)_i]} \\
     & \le \prod_{i=1}^n \exp(\frac{\Delta_i}{s}) = \exp(\sum_{i=1}^n \frac{\Delta_i}{s}) = \exp(\frac{\sum_{i=1}^n \Delta_i}{s}) = \exp (\epsilon).
    \end{split}
    \end{equation}
    Note that $\sum_{i=1}^n \Delta_i = \sum_{i=1}^n |f(x)_i - f(y)_i|$ is the 1-norm of the vector $f(x)-f(y)$.
\end{proof}

\subsection{Proof on Proposition \ref{prop:epsilon}}
\begin{proof}
    Let $D$ and $D'$ be adjacent datasets, $\mathcal{A_H}$ be the machine learning algorithm, $M_{AL}: \mathbb{R}^n \longrightarrow \mathbb{R}^n$ be our additive logistic mechanism with parameter $s$. We train the algorithm $\mathcal{A_H}$ using $D$ and $D'$ respectively to obtain two models $NN_{\theta}: \mathcal{X} \longrightarrow \mathbb{R}^n$ and $NN_{\theta'}: \mathcal{X} \longrightarrow \mathbb{R}^n$, with parameters $\theta = \mathcal{A_H}(D)$ and $\theta' = \mathcal{A_H}(D')$.
    
    We migrate Equation \ref{proof:dp2} to this scenario and get
    \begin{equation}
        \frac{\mathrm{P}[M_{AL} (x, \mathcal{A_H}, s) = \Vec{z}]}{\mathrm{P}[M_{AL} (y, \mathcal{A_H}, s) = \Vec{z}]} 
        = \prod_{i=1}^n \frac{\mathrm{P}[Log(s)_i = \Vec{z}_i - \mathcal{A_H}(x)_i]}{\mathrm{P}[Log(s)_i = \Vec{z}_i - \mathcal{A_H}(y)_i]}
        \le \exp(\sum_{i=1}^n \frac{\Delta_i}{s}) = \exp (\epsilon).
    \end{equation}
    We have shown that $\mathrm{P}[M_{AL} (x, \mathcal{A_H}, s) = \Vec{z}] \le \exp (\epsilon) \cdot \mathrm{P}[M_{AL} (y, \mathcal{A_H}, s)]$, by Definition \ref{prop:DP}, we conclude that the machine learning algorithm $\mathcal{A_H}$ is $\frac{\Delta_{\mathcal{H}}}{s}$-differentially private.
    
    From Remark \ref{remark:1-norm} we know that $\Delta_{\mathcal{H}} = \sum_{i=1}^n |\mathcal{A_H}(x)_i - \mathcal{A_H}(y)_i| = \sum_{i=1}^n \Delta_i$ is the sensitivity of $\mathcal{A_H}$ that computed in 1-norm distance.
\end{proof}

\end{document}